# An Automated Robotic Arm: A Machine Learning Approach

Krishnaraj Rao N S
*Department of CSE*
*St. Joseph Engineering College*
Vamanjoor, Karnataka, India
krisndi@gmail.com

Avinash N J
*Department of ECE*
*New Horizon College of Engineering*
Bangalore, Karnataka, India
avinash.yuvaraj@gmail.com

Rama Moorthy H
*Department of CSE*
*St. Joseph Engineering College*
Vamanjoor, Karnataka, India
ramamoorthy.h@ieee.org

Karthik K
*Department of CSE*
*St. Joseph Engineering College*
Vamanjoor, Karnataka, India
2karthik.bhat@gmail.com

Sudesh Rao
*Department of AIML*
*N.M.A.M.I.T, Nitte*
Karnataka, India
sudesh.rao@nitte.edu.in

Santosh S
*Department of ISE*
*N.M.A.M.I.T, Nitte*
Karnataka, India
santhosh.s@nitte.edu.in

*Abstract*—The term 'robot' generally refers to a machine that looks and works in a way similar to a human. The modern industry is rapidly shifting from manual control of systems to automation, in order to increase productivity and to deliver quality products. Computer-based systems, though feasible for improving quality and productivity, are inflexible to work with, and the cost of such systems is significantly high. This led to the swift adoption of automated systems to perform industrial tasks. One such task of industrial significance is of picking and placing objects from one place to another. The implementation of automation in pick and place tasks helps to improve efficiency of system and also the performance. In this paper, we propose to demonstrate the designing and working of an automated robotic arm with the Machine Learning approach. The work uses Machine Learning approach for object identification / detection and traversal, which is adopted with Tensorflow package for better and accurate results.

## I. Introduction

A 'robot' is a system that can operate in a way similar to human actions, i.e., a system that takes decisions and accomplishes tasks assigned to it without external intervention. The development of the idea of robots led to the birth of the field of robotics, and robotics, in turn, led to the emergence of fields like automation. Industrial automation, though implemented already, requires more tuning in terms of reducing human intervention in the working of the automated systems. This is where concepts like machine learning and object detection come into picture [1]. Machine learning is a category of artificial intelligence that deals with formulation of models and procedures in order to implement the concept of pseudo-intelligence in systems. Machine learning makes use of networks and algorithms to design a model that can tackle the provided problem statement and lead to the creation of an intelligent systems. The main objective of machine learning is to enable systems to learn from previous steps and data examples without the need for being programmed for every future task or action to be performed.

It enables the implementation of pattern realization, classification, object detection, and many other concepts, that has led to the development and flexibility of many useful fields and areas like medical diagnosis, weather prediction, biometrics, spam identification, etc. Object detection deals with the identification and distinction of objects into different categories. The real-world objects can be captured either as image or a live streaming video through a camera module, and then can be processed using object detection codes to segregate the object into groups like bottle, tree, bird, cat, box, etc. The working of object detection can be considered identical to the working of human eye. The human eye can capture and classify thousands of real-world objects in fraction of seconds. With high speed GPU and faster algorithms, the same functionality can be achieved using camera module and a computer system. Object detection finds wide range of use in areas like image identification and face detection used for bio-metrics, video surveillance, etc.

## II. Literature Survey

There have been many attempts in the design of robots and end effectors that can be used for industrial purposes. Animate, a simple robotic arm implemented for automation of die-casting in the factory was the first industrial robot developed in 1961 [2].

In another approach, Abhinav Sinha et. al have put forward a mathematical model for the control of pick and place manufacturing robots employing inverted pendulum concept and have proven it mathematically. But no real-time implementation is carried out to prove its efficiency [3].

Vonasek et al. [4] proposed an approach for locomotion planning of robotic operators for a bin-picking industrial robot. The planner is called RRTTS-MP (RRT with Task-Space Motion Primitives). They have tested this planner in simulation and also as a prototype. The RRT-TS-MP method surpasses existing TS-RRT method in many ways. But the main disadvantage is its accuracy and computational time. A robotic gripper with modifiable stiffness using shape memory alloy (SMA) and a model for grasping force is proposed by Yin et al. [5]. It is verified using a two jawed robotic gripper. The stiffness of gripper varied using the relation between temperature and elastic modulus of shape-memory alloy wires. Results reveal that the grasping capability increases by at least 30% of normal two fingered grippers but the drawback is, it is applicable for only small weights.

Gauchel and Schell [6] has proposed a model individually movable two jawed pneumatic gripper using closed-loop position controller of jaws. But in pneumatic based approach, cost and energy consumption is high.

Glick et al. [7] have designed a robotic gripper with a combination of fluidic and elastomeric actuators and gecko-inspired adhesives. This has increased the soft gripping property. But the disadvantage of the mentioned technique is position and speed of gripping is complex to control and needs more computation for lifting heavy parts.

Omar [8] presented a pick and place automated robotic arm constrained by PC vision. The robot could pick the article in a particular position only. The robot used a mechanical gripper. Along these lines, it cannot deal with the item securely. Articles in the specified location are lifted by the automated arm.

Ronanki and Kranthi [9] presented "Design and Fabrication of Pick and Place Robot to be used in Library". This robot is meant to get books from the library and conveys them to the goal. The automated arm utilized here can deal with items in any location. RFID labels are utilized to recognize the books. This framework can do this particular assignment only and it's a line following robot. Each RFID has its very own way, and this makes the framework progressively intricate.

Begum and Vignesh [10] have planned an independent robot con- figuration controlled using Android utilizing remote vitality. Here the framework works as indicated by voice directions or discourse conveyed by the client and the automated arm is suitable for getting hold of the objects of any kind and in any position. RF innovation is utilized so that viewable pathway is a noteworthy imperative in correspondence.

Yoshimi et al. [11] presented a framework for getting task of items of less width using automated arm having gripper that is two-fingered and parallel. Less thick items like paper and cards are grabbed using this mechanical arm. The articles fall down because of use of a parallel gripper [12].

Baby et al. [13] in "Pick and Place Robotic Arm Implementation Using Arduino" proposed design of an arm controlled with Arduino to pick and place objects through user-specified commands. The robot is controlled using an Android device such as a smartphone using Bluetooth.

## III. PROPOSED MEHTODOLOGY

The proposed system aims to implement the prototype of an automatic robotic arm that picks and places objects. This system focuses on the implementation of the concepts of machine learning for deciding the amount of movement that the arm must produce in order to reach and grab the target object.

### A. System Design

The robotic arm is made up of PLA material and is 3D printed. The arm consists of 5 servo motors, one for each joint of the arm as shown in Fig. 1. There are one MG996, two MG995 and two MG90S motors. These motors are controlled via the Raspberry Pi controller, which is remotely controlled using a headless display via VNC viewer. Python programs for object detection and arm movements is fed into the controller. The camera module captures image or video as the input to the object detection codes, and the object to be picked is identified. Then, rotations for each motor, in order to pick up the object, are determined and are given to the motors in terms of duty cycle via the GPIO pins. The end point of the arm called end effector or gripper is used to grab the objects.

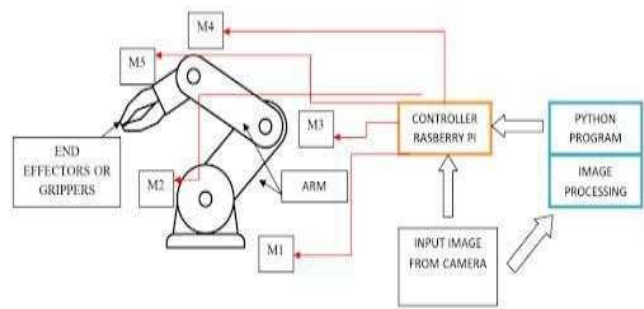

Fig. 1. System design block diagram of arm.

### B. Hardware Components

In this section, we will present the hardware components utilized to design and develop the robotic arm as displayed in the Fig. 2 [14].

*1) Robotic Arm:* A robotic arm (Fig. 2) is a sort of arm which may be mechanical, and can be programmed with capabilities almost equivalent to an actual arm. The robotic arm can be aggregate of a component or a piece of a more advanced robot. The connectivity of these controllers is integrated by joints that either allow rotational movement, or displacement. The connections of the controllers are supposed to make a kinematic chain. The endpoint of the controller where the chain ends is called the end effector. End effector is similar to actual human hand.

*2) End effectors:* In mechanical autonomy, end effector (Fig. 2) is a gadget towards the finish of the arm, intended for collaborating with surroundings. The correct idea of gadget relies upon utilization of arm. In other words, end effector implies the final connection of automaton. Tools are fixed at this end most position. In more extensive meaning, end effector is viewed as a segment of automaton which connects to a workplace. Thus said, it does not hint at wheel of the movable automaton that is likewise not end effector—it is a portion of the automaton's portability. The purpose of the end effector is to absorb the ideal task assigned to it, for example, grasping, welding, etc. For instance, in automotive assemblies, robotic arms can perform a bunch of errands like welding and revolution and positioning of parts at the time of assemblage, in a few conditions where close resemblance of the human arm is needed, as in robots intended to conduct bomb demobilization and transfer, etc. End effector may comprise of gripper or an apparatus. While alluding to automaton control, we have four general grippers:

*a) Impactive:* jaws or hooks which can bodily clutch the object by straight effect on the article.

*b) Ingressive:* sticks or needles which bodily enter the outside of any article.

*c) Astrictive:* attractive exertions appertained to the object's surface (regardless of environment).

*d) Contigutive:* require physical contact for gripping (for example, glue, surface tension or solidifying).

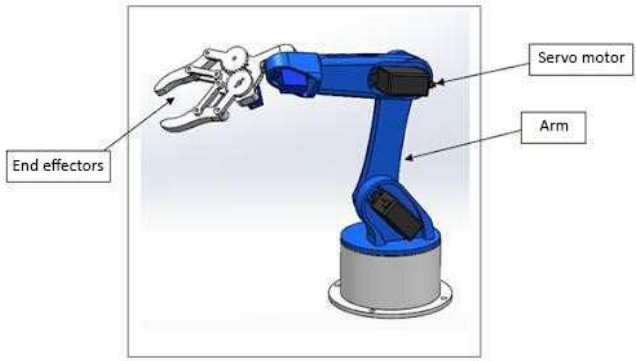

Fig. 2.  3D model made in SOLIDWORKS.

*3) Servo motor:* Servo motors ( Fig. 3.) are rotary or linear actuators which provide control over slant or level position, speed and acceleration. They consist an engine coupled with a sensor for input of position. They additionally require moderately refined controller, often, a committed component planned explicitly to be used with the motors. Servo motors do not belong to a class of engines, in spite the term servo motor being frequently used for indicating to an engine convenient for utilization in a closed-circle control framework. Servo motors have applications in robotics, automated fabrication, CNC machinery, etc.

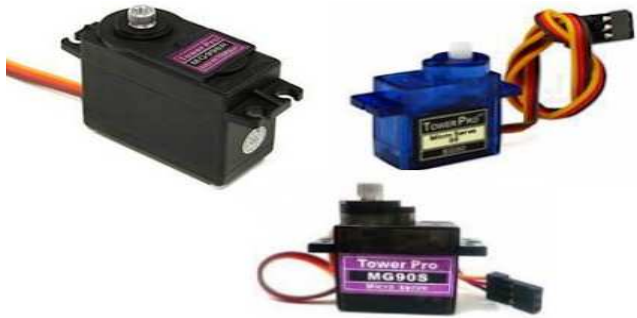

Fig. 3.  Servo motors used in the arm.

*4) Raspberry Pi:* Unlike the well-known standard boards like Arduino, Raspberry Pi is a real PC with Operating System (OS) such as Windows in PC and Apple iOS in MacBook. This enables Raspberry Pi to do considerably more entangled employments than Arduino can do. It is an efficient gadget that permits individuals of any age to review computations, and to figure out how to code in dialects like Scratch and Python.

*5) GPIO:* A 40-stick GPIO header is present on all Raspberry Pi 3B+ model (Fig. 4). These pins (also called GPIO pins) serve various purposes. There are pins for power supply (3.3V at pins 1 and 17, and 5V at pins 2 and 4), signal out and ground connections (6, 9, 14, 20, 25, 30, 34 and 39). These pins can be used in one of the two ways: using the BCM (Broadcom channel) number and its BOARD number. Either of these modes can be chosen in Python:

  *a)* GPIO.setmode(GPIO.BCM) Or
  *b)* GPIO.setmode(GPIO.BOARD)

Fig. 4.  Pin configuration of Raspberry Pi.

*6) Ultrasonic sensor:* Ultrasonic sensor is a sensor that measures the distance to an article using ultrasonic sound waves. It uses transducer to transmit and receive ultrasonic pulses that transfer the data from the item's vicinity. High-recurrence sound pulses reflect from the object to deliver particular reverberation patterns. The ultrasonic sensor used is the HC-SR04 sensor. It has four pins: ground pin (GND), trigger input (TRIG), power supply of 5V (Vcc), echo pulse output (ECHO). The module can be controlled with Vcc, grounded via GND. Raspberry Pi can be used to send information flag via the TRIG pin, that would trigger the sensor to transmit ultrasonic pulses. These transmitted waves skip off the adjacent items. Some ultrasonic pulses will be bounced back towards the sensor. The arrival pulses are identified by the sensor, the time between the trigger and received wave is calculated, and then 5V motion is transmitted through the ECHO pin.

*7) Camera module:* The Raspicam camera module is used in this work for viewing and tracking the object's position. The camera serves as the eyes of the robotic arm, and permits the arm to "see" on its own and locate the object. The Raspicam can be used to stream videos, and to capture static photographs of objects. The installation of the camera module over the Raspbian involves a sequence of steps:

  *a)* Insert the camera into the slot provided between the HDMI port and the Ethernet port.
  *b)* Go to Raspberry pi configuration settings and enable the camera interface, reboot.
  *c)* After reboot, get the most recent Raspberry pi firmware by executing:
  sudo apt-get update or sudo apt-get upgrade
  *d)* The camera is enabled from the Raspberry pi configura- tion program using the command:
  sudo raspi-config

The Raspicam is now configured for use, and can be used to capture photos or record videos by running the respective programs or commands.

*8) Barometric pressure sensor:* Barometric pressure sensor is a computerized pressure sensor module based on BMP-180. It utilizes piezo-resistive innovation for accuracy, linearity, EMC robustness and steadiness for a more extended time-frame. It has a control unit with E2PROM and a sequential I2C interface and an analog to digital converter. It can be used to determine the change in pressure on a surface. In the work, this sensor is used for determining if the gripper is holding on to the object once it has been picked up. This helps in improving the accuracy of picking.

*C. Software Components*

Here in this section, we will look into the various software components/Tools used in our work:

*1) Raspbian:* Raspbian is an operating system for Raspberry Pi that is Debian-based. It makes use of Pi Improved X-Window Environment [1] (PIXEL). It is made up of an improvised LDXE desktop environment, and the Openbox stacking window manager-it has new settings along with several alterations. This version comes along with a computer algebra software Mathematica and Minecraft Pi , a different make of Minecraft [15].

*2) TensorFlow:* TensorFlow is an open-source library developed by Google that is useful for mathematical computations and helps in the easy implementation of machine learning applications. TensorFlow provides front-end API through Python programming language. TensorFlow is useful for systems using machine learning as it helps in the creation of models and graphs, where the nodes are basically Python objects. The computations required to be performed are done using C++, which proves to be beneficial in terms of performance and time. TensorFlow's changeable design allows plain arrangement of calculations over several categories (CPUs, GPUs, TPUs), task areas and server clusters, to transferable gadgets. TensorFlow was initially created by specialists and designers from the Google's AI association. Its purpose is to offer the abstraction for the implementation of machine learning and allows the developers to concentrate mainly only the application logic, instead of worrying about the complex computations.

*3) OpenCV:* OpenCV is an open-source library that provides algorithms for implementing computer vision and serves as an image processing library for machine learning applica- tions. OpenCV algorithms can be used for face recognition, live tracking, identify object locations, process images, etc. OpenCV can take video as input from a camera, or through a video file [16]. For this, a VideoCapture object needs to be created as:

obj = cv2.VideoCapture(argument)

where 'argument' is the index value of the camera connected to the system for live video capture, and is the name of the video file for stored video.

*4) PuTTY software:* PuTTY runs by dispatching written commands and receives some text acknowledgements through a TCP/IP socket in the same way as conventional terminals (TTY). But the difference is that PuTTY makes use of secure socket (SSH) where the packet payloads are wrapped using public key encryption. PuTTY freely implements SSH and Telnet for Unix as well as Windows operating systems and it also has an xterm terminal emulator. PuTTY serves like a substitute for clients making use of telnet. The main benefit of PuTTY is the SSH dispenses safe, encoded link to isolated systems. It is fully-featured, coherent and its terminal emulator is acceptable.

*5) VNC viewer:* Virtual network computing (VNC) is a graphical system with desktop-sharing. It can be used to control another computer which is located in some other place via the Remote Frame Buffer protocol (RFB). It receives and transfers mouse and keyboard interrupts from system to system. While doing so, it passes the graphical-screen updates backwards through the Web. VNC is OS-independent. VNC has servers and users for several GUI-based platforms and Java. Many numbers of users can simultaneously be connected to one VNC server.

*6) Python:* Python is an interpreted, simple and general-purpose programming language. Python provides code syntax that are quite easy to learn and implement, and hence is a good choice for beginners. Python is also developer-friendly, as it provides functionalities similar to any other programming language. Python provides an IDLE for scripting, which en- ables developers to write and edit code in the editor and run it in the Python shell. This IDLE provides features such as autocomplete suggestions, automatic indentation, interpreter, packages, etc. The IDLE also provides an interface for the interpreter, where the user can run Python commands directly in the shell.

*D. Architecture Adopted*

A simple architecture is proposed, following flow diagram (Fig. 5) depicts the architecture used:

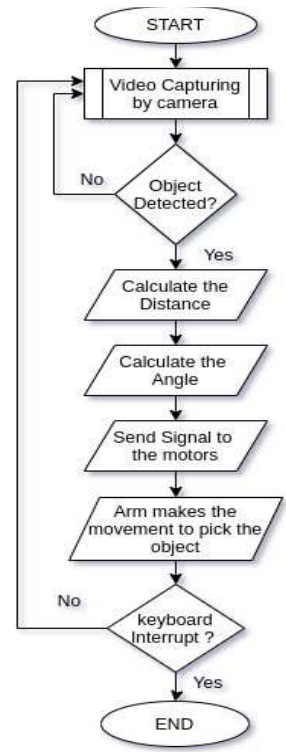

Fig. 5. Flow architecture Proposed.

The system prototype accomplishes the task of picking and placing objects as expected. The robotic arm works as follows:

The controller, Raspberry Pi, has the Raspbian OS installed onto it. This OS is accessed via the VNC viewer software, which is helpful for remote controlling of devices in the same network for a headless display.

The camera module, Raspicam, is used to view the object to be picked. The object must be present in the scope of the camera. The object detection code written using OpenCV and TensorFlow libraries enables real-time detection and identification of the object. In our work, the object detection code takes real-time video as input, and calculates the percentage of similarity. Once the object is identified then only then the arm will pick it.

*1) Distance Calculation:* Let H be the width of the object, F be the focal length of the camera, and h be the apparent height. Distance D is calculated as in equation (1):

$$D = (W * F) / h \qquad (1)$$

This distance D is the distance of the object from the camera. The distance of the object from the arm is calculated using the following code. This code also calculates the movements to be made by each motor in order to reach the object. The related code is:

position_to_voltage (x, y, z):

a = 14.9; // length of robot segment n°1

b = 14.6; // length of robot segment n°2

c = 5.4; // length of robot segment n°3

r = math.sqrt(x2 + y2 + math.pow((z-c),2));

u = (a2 - b2 + r2) / (2 * r);

*2) Angle Estimation:* Now that we know the distance from the object to the arm, we need to know the length of the two arm segments. This is represented in the below figure(Fig. 6):

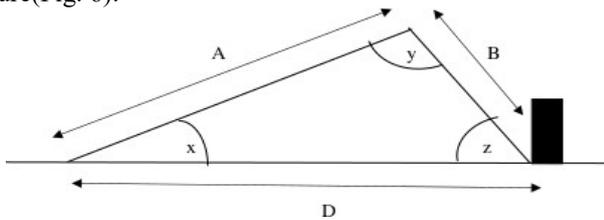

Fig. 6. Angle Estimation.

A and B are the two arm segment widths. Angles are written in the code as:

x = arccos((A 2 + D 2 – B 2 ) / 2AD)

y = arccos((A 2 + B 2 – D 2 ) / 2AB)

z = arccos((B 2 + D 2 – A 2 ) / 2BD)

The above, are written in the structure of the program as following:

// angle conversions

g = math.atan((z-c)/math.sqrt((x2 + y2)));

0 = math.atan (-x/y);

1 = math.acos (math.sqrt(a2 – u2) /a) -g;

2 = math.acos (- (a2 – u2) /b) -t1 -g;

// voltage conversions

angle [0] = 0.02 + 3.3 * 0;

angle [1] = -2.4 + 3.3*(math.PI/2 - 1);

angle [2] = -10 + 3.3*(math.PI-2);

angle [3] = (1 + 2 – math.PI) / 0.3;

*3) Converting Angles into Duty Cycle:* Each motor on the arm has 3 connections to be made-signal, power supply and ground. The signal wire is connected to the GPIO pins of the Raspberry Pi in order to send signals to each motor as PWM waves. To do this, the duty cycle for the corresponding angles is calculated as in equation (2):

$$duty = angle / 180 + 2 \qquad (2)$$

where 'angle' is the desired angle for the motor rotation. Then, this value of duty cycle is given as input to the GPIO pin of the corresponding motor as follows:

pwm = GPIO.PWM(pinnumber, GPIO.OUT )

pwm.ChangeDutyCycle(duty)

The motors make movements according to the duty cycle and pick and place the object from one position to the specified home position.

## IV. RESULTS

The following figures depicts the proposed model identifying the object (Fig. 7) with the similarity ratio, finding the distance and calculating angle inclined to the object and arm. The arm then picks it up (Fig. 8).

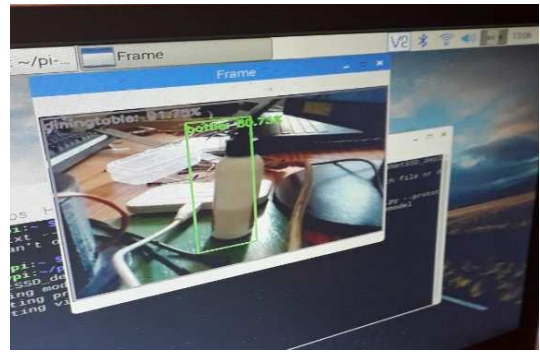

Fig. 7. Identifying the Object.

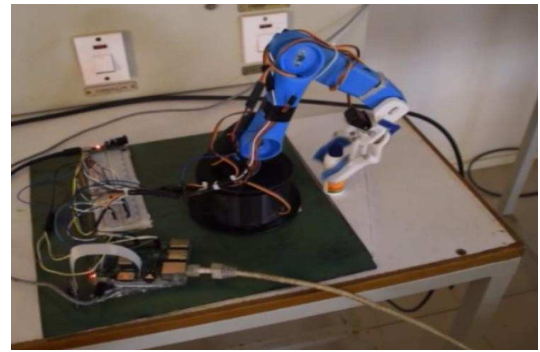

Fig. 8. Picking up the object.

## V. CONCLUSION

The robotic arm that is implemented using Raspberry Pi, uses branches of machine learning to pick and place specified objects and aims at avoiding human intervention in the industries where automation can be achieved in a very efficient manner. In today's situation, wherein man power and time are the critical factors for the accomplishment of tasks, automation can help to save a majority of the investments that are to be dedicated towards labour and retardation in the work completion when not using automation can be avoided. The robotic arm prototype focuses on demonstrating the pick and place characteristic of robots in which case, objects are to be carried from one location to another, which is mainly required for industrial automation. One of the applications of this robotic arm is to pick and place parts, from a conveyor line and place it accurately in a particular position or on the next processing unit for further manufacturing procedures. The usage of machine learning can perform a lean manufacturing by eliminating unnecessary steps involved in performing the task. This will result improvement of manufacturing lead time to produce a particular product.